\documentclass[pdflatex,sn-mathphys-ay,iicol]{sn-jnl}

\usepackage{graphicx}
\usepackage{multirow}
\usepackage{amsmath,amssymb,amsfonts}
\usepackage{amsthm}
\usepackage{mathrsfs}
\usepackage[title]{appendix}
\usepackage{xcolor}
\usepackage{textcomp}
\usepackage{manyfoot}
\usepackage{booktabs}
\usepackage{algorithm}
\usepackage{algorithmicx}
\usepackage{algpseudocode}
\usepackage{listings}
\usepackage{array}
\usepackage{textcomp}
\usepackage{stfloats}
\usepackage{url}
\usepackage{verbatim}
\usepackage{cite}
\usepackage{booktabs}
\usepackage{color}
\usepackage{nicematrix}
\usepackage{threeparttable}
\usepackage{makecell}
\usepackage{bbding}
\usepackage{ulem}
\usepackage{hhline}
\usepackage{bm}
\usepackage{wrapfig}
\usepackage{caption}
\usepackage{amsthm}
\usepackage{listings}
\usepackage{threeparttablex}

\theoremstyle{thmstyleone}

\theoremstyle{thmstyletwo}

\theoremstyle{thmstylethree}

\newcolumntype{L}[1]{>{\raggedright\let\newline\\\arraybackslash\hspace{0pt}}m{#1}}
\newcolumntype{C}[1]{>{\centering\let\newline\\\arraybackslash\hspace{0pt}}m{#1}}
\newcolumntype{R}[1]{>{\raggedleft\let\newline\\\arraybackslash\hspace{0pt}}m{#1}}

\definecolor{myBlue}{HTML}{3ba9db}
\definecolor{myPink}{HTML}{E7DBCC}
\definecolor{myBlue}{HTML}{5999B6}
\definecolor{myYellow}{HTML}{F1F4AE}
\definecolor{myHighlight}{HTML}{0000ff}

\usepackage{etoolbox}
\makeatletter
\AfterEndEnvironment{algorithm}{\let\@algcomment\relax}
\AtEndEnvironment{algorithm}{\kern2pt\hrule\relax\vskip3pt\@algcomment}
\let\@algcomment\relax
\newcommand\algcomment[1]{\def\@algcomment{\footnotesize#1}}
\renewcommand\fs@ruled{\def\@fs@cfont{\bfseries}\let\@fs@capt\floatc@ruled
  \def\@fs@pre{\hrule height.8pt depth0pt \kern2pt}
  \def\@fs@post{}
  \def\@fs@mid{\kern2pt\hrule\kern2pt}
  \let\@fs@iftopcapt\iftrue}
\makeatother

\begin{document}

\newtheorem{lem}{Lemma}
\newtheorem{prop}{Proposition}

\title[Article Title]{De-biasing Skeleton-based Action Recognition with Convex Hull Adaptive Shift}

\author[1]{\fnm{Mengyuan} \sur{Liu}}
\equalcont{These authors contributed equally to this work.}

\author*[2]{\fnm{Yuhang} \sur{Wen}}\email{wenyh29@mail2.sysu.edu.cn}
\equalcont{These authors contributed equally to this work.}

\author[3]{\fnm{Yi} \sur{Zhang}}

\author[4]{\fnm{Songtao} \sur{Wu}}

\author[1]{\fnm{Hong} \sur{Liu}}

\author[5]{\fnm{Junsong} \sur{Yuan}}

\author*[6]{\fnm{Beichen} \sur{Ding}}\email{dingbch@mail.sysu.edu.cn}

\affil[1]{\orgdiv{State Key Laboratory of General Artificial Intelligence}, \orgname{Peking University, Shenzhen Graduate School}, \orgaddress{\country{China}}}

\affil[2]{\orgdiv{School of Intelligent Systems Engineering}, \orgname{Sun Yat-sen University}, \orgaddress{\country{China}}}

\affil[3]{\orgname{City University of Hong Kong}}

\affil[4]{\orgdiv{R\&D Center}, \orgname{Sony China Ltd.}, \orgaddress{\country{China}}}

\affil[5]{\orgname{University at Buffalo SUNY}, \orgaddress{\country{USA}}}

\affil[6]{\orgdiv{School of Advanced Manufacturing \& Southern Marine Science and Engineering Guangdong Laboratory (Zhuhai)}, \orgname{Sun Yat-sen University}, \orgaddress{\country{China}}}

\abstract{Skeleton sequences can represent both individual actions and multi-entity interactions, encompassing human bodies, hands, objects, and robots. Existing approaches to recognize skeleton-based actions and interactions usually adopt a late fusion strategy, which expects individuals are independent and identically distributed to train a robust weight-shared entity encoder. However, observed entity bias in various skeletal data violates this assumption, leading to suboptimal optimization of backbone models that might produce wrong recognition results. This bias arises from the world coordinate system's initial configuration, where the choice of origin often creates bias in representation. To this end, we propose a Convex Hull Adaptive Shift based normalization method to reduce Entity bias (CHASE), improving performance across a variety of skeleton-based action and interaction recognition tasks. To adaptively apply plausible shifts to the input skeletons, we formulate a plug-and-play parameterized network that ensures the relocated world origin lies within the skeleton convex hull, which avoids non-convergence by limiting the search space. To further minimize entity bias, we incorporate an auxiliary objective that leverages pair-wise distribution distances to guide network optimization. To support both single- and multi-entity actions, we propose a sub-entity strategy that offers a consistent formulation for both scenarios. Moreover, CHASE demonstrates compatibility with various intra-skeleton modalities, such as bones and velocities, highlighting its adaptability. Essentially, our method works as a normalization approach to reduce entity bias, enabling subsequent classifiers to achieve improved recognition performance across diverse settings. Extensive experiments on seven datasets, including NTU RGB+D, NTU RGB+D 120, H2O, Assembly101, Collective Activity, Volleyball, and HARPER, consistently verify our approach by seamlessly integrating with various backbones and significantly boosting their performance. Our code is publicly available at~\url{https://github.com/Necolizer/CHASE}.}


\maketitle

\section{Introduction}\label{sec:intro}
Skeleton-based action recognition has been studied extensively for individual subjects, and many well-performing models exist for single-person action datasets \citep{pivit2024CVPR,Rajasegaran_2023_CVPR,9975251}. Extending these models to multi-entity settings is more difficult, as interactions can occur between human bodies \citep{NW-UCLA,liu2017pkummd}, hands \citep{Assembly101,Ohkawa_2023_CVPR}, objects \citep{H2O_TA-GCN2021,Garcia-Hernando_2018_CVPR}, and robots \citep{HARPER2024,chico2022}, each requiring the model to jointly reason over multiple entities with different roles. This breadth of scenarios supports applications in scene understanding \citep{ren2024spikepoint,jiang2024scaling,Plizzari2024-sp}, human foundation models \citep{Tang_2023_CVPR,Ci_2023_CVPR,khirodkar2024_sapiens,wang2025foundation}, and human-robot collaboration \citep{Du2024Constrained,jahangard2024jrdbsocial}. Skeletal representations offer a compact and appearance-invariant encoding of spatiotemporal pose \citep{yang2024skeleton,Wanyan2025,yang2024view}, which makes them a natural choice for this task. This paper addresses a data-level problem that has so far received little attention in the literature: the raw skeletal data for multi-entity actions carries inherent distributional asymmetries between entities, and these asymmetries systematically degrade the optimization of standard backbone models before any architectural choice is made.

The predominant strategy for handling multiple entities in skeleton-based recognition is late fusion \citep{CTR-GCN2021,InfoGCN2022,hdgcn2023,dstanet2020,STSA-Net2023,degcn2024tip}. A weight-shared backbone independently encodes each entity, and the resulting feature vectors are averaged to produce the final representation. This approach is convenient because it reuses existing single-entity models without modification. It rests on the assumption that all entities are independently and identically distributed (IID), so that a single shared encoder can be expected to perform equally well on each of them. As we show in this paper, this assumption is frequently and substantially violated in practice.

\begin{figure}[t]
    \begin{center}
    \includegraphics[width=\linewidth]{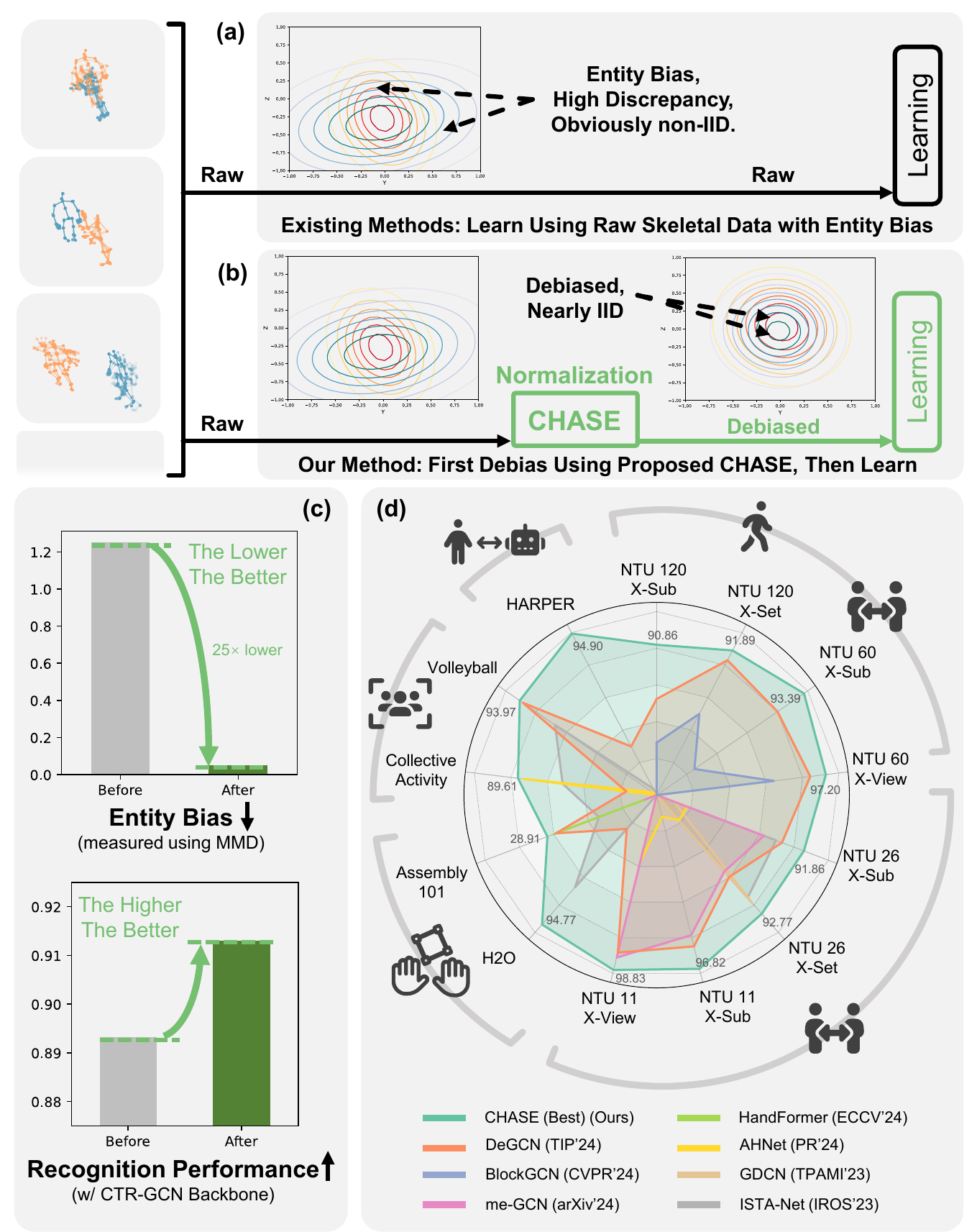}   
    \end{center}
    \vspace{-0.5em}
    \caption{\textbf{Overview.} (a) Existing methods train their models on raw skeletal data with inherent entity bias, resulting in unsatisfactory performance. (b) Our proposed CHASE works as a normalization method to de-bias the skeleton sequences. (c) It can significantly reduce the inter-entity distribution discrepancy and improve recognition performance. (d) CHASE help backbones achieve state-of-the-art results across a variety of action and interaction learning tasks.}
    \label{fig:intro}
    \vspace{-1.0em}
\end{figure}

We identify a fundamental issue in skeletal data for multi-entity actions, which we term \textit{Entity Bias}. Entity bias refers to significant distributional discrepancies between skeleton sequences of individuals with different entity indices, a characteristic that is particularly evident in raw skeleton data. This bias originates from the initial configuration of the world coordinate system, where the choice of origin introduces inherent asymmetries into the representation. A representative example is the standard preprocessing of NTU RGB+D 120 \citep{NTU120}, one of the most widely used benchmarks. The coordinate origin is conventionally placed at the spine-base joint (pelvis) of the first person, a choice motivated by its stability compared to extremity joints. This convention works well for single-entity recognition. In multi-entity settings, however, it systematically favors the first entity: the first person is always centered at the origin while subsequent persons are spatially displaced, and their skeletal distributions diverge by amounts proportional to their physical separation from the reference entity. More broadly, any fixed choice of reference joint or reference entity during data collection or preprocessing will produce the same kind of asymmetry. As illustrated in Fig.~\ref{fig:intro} (a), entities with different indices are visualized in blue and orange, with their respective distributions (based on \(10^4\) mutual action samples from NTU120 \citep{NTU120}) projected onto a 2D plane. The results reveal notable discrepancies in both the mean and covariance of these two entity distributions. The late fusion strategy relies on the IID assumption to train a robust weight-shared classifier, and entity bias directly violates this assumption, resulting in suboptimal recognition performance. This raises the question: \uline{Is there a way to mitigate entity bias in skeleton sequences?}

To address the above problem, we propose CHASE, a Convex Hull Adaptive Shift based normalization method that reduces entity bias across a wide range of skeleton-based action and interaction recognition tasks. The core idea is to \uline{adaptively relocate skeleton sequences so that their per-entity distributions become approximately IID}, as illustrated in Fig.~\ref{fig:intro} (b). CHASE consists of four components. A learnable plug-and-play module applies sample-adaptive shifts to the input skeletons, with the relocated coordinate origin constrained to lie within the skeleton convex hull. An auxiliary objective based on pairwise distribution distances guides the network toward distributional alignment during training. A sub-entity strategy extends CHASE to single-entity settings by decomposing one entity into structurally comparable sub-entities, so the same distribution alignment objective applies consistently. A modality-agnostic formulation ensures that CHASE operates on any skeletal representation without requiring modality-specific design choices. In essence, CHASE works as a normalization step that produces de-biased, nearly IID skeleton samples, improving the optimization of subsequent weight-shared backbone models. As shown in Fig.~\ref{fig:intro} (c)(d), experiments on 7 datasets confirm consistent improvements across diverse interaction types and backbone architectures.

Our main contributions are three-fold:
\begin{itemize}
    \item The observed entity bias in skeleton sequences undermines the effectiveness of weight-shared backbones and results in suboptimal performance. To the best of our knowledge, we are the first to investigate the issue of the entity bias. Our main idea is to address this bias by adaptively re-centering the coordinate system for each sample, thereby significantly improving the performance of the subsequent backbone models.
    \item We proposed CHASE, a novel de-biasing method based on convex hull adaptive shift, serving as an additional normalization step for the subsequent backbone. Specifically, CHASE consists of a learnable network and an auxiliary objective, which learns origin offsets within the skeleton convex hull and guides the bias minimization.
    \item Extensive experiments on 7 diverse datasets consistently verify our proposed method by improving single-entity backbones to achieve state-of-the-art performance across a variety of recognition tasks.
\end{itemize}

This paper is an extension of our conference paper \citep{wen2024chase}. The new contributions of this paper include:
\begin{itemize}
    \item In \citep{wen2024chase}, we formulated CHASE for multi-entity actions. Compared to the previous work, in this paper, we propose the Sub-Entity Strategy to extend CHASE to any-entity skeleton sequences, which includes both single- and multi-entity actions. This strategy offers a unified approach by conceptualizing a single entity as a collection of sub-entities derived from its components. Specifically, we define the criteria for selecting sub-entities, incorporating both subgraph and isomorphism conditions. Additionally, we illustrate through examples how sub-entities can be identified using heuristic methods, such as leveraging the physical topology of the entity graph in the skeleton sequences.
    \item In \citep{wen2024chase}, we only applied CHASE to joint modalities. Compared to the previous work, in this paper, we extend CHASE to accommodate diverse widely-used skeletal representations, including joints, bones, velocities, and JBF \citep{degcn2024tip}. This extension involves reformulating the learnable network, auxiliary objectives, and the sub-entity strategy for these modalities, thereby demonstrating the adaptability of CHASE in reducing bias and improving performance across diverse intra-skeleton modalities.
    \item In \citep{wen2024chase}, we conducted experiments with 4 backbones and 6 benchmarks, including NTU RGB+D \citep{NTU60}, NTU RGB+D 120 \citep{NTU120}, H2O \citep{H2O_TA-GCN2021}, Assembly101 \citep{Assembly101}, Collective Activity \citep{cad2009}, and Volleyball \citep{msibrahi2016VOL}. Compared to the previous work, in this paper, we evaluate our method with more recently-proposed baseline backbones and also in genuinely new scenarios. In particular, we include HARPER \citep{HARPER2024}, a human-robot interaction benchmark that introduces a fundamentally different skeleton topology: human and quadruped robot skeletons have distinct joint structures and dramatically different morphologies, making the entity bias both more severe and structurally heterogeneous than the human-human settings in prior work. CHASE achieves a significant improvement on HARPER, demonstrating its applicability in this challenging new domain where existing approaches are especially ill-suited. In addition, we provide extensive comparisons against more state-of-the-art methods. Further experiments are conducted to show that the proposed method can be effectively applied to single-entity actions and various skeletal representations. Notably, in NTU RGB+D and NTU RGB+D 120 benchmarks, CHASE significantly improves the ensemble accuracy of the state-of-the-art DeGCN \citep{degcn2024tip}.
\end{itemize}

\section{Related Work}\label{sec:relatedwork}

\subsection{Distribution Alignment and Normalization}

Reducing distributional discrepancies between different subpopulations or domains is a long-studied problem. Domain adaptation methods align feature distributions across domains through statistical divergence minimization \citep{long2015dan} or adversarial training \citep{ganin2016domain}. These methods are designed for cross-domain settings where the shift occurs between distinct datasets or collection conditions and where domain labels are explicitly available.

Normalization layers address a complementary problem: stabilizing optimization by reducing internal covariate shift during training. Batch Normalization \citep{ioffe2015batch} standardizes activations over each mini-batch. Instance Normalization \citep{ulyanov2016instance} operates per sample, Layer Normalization \citep{ba2016layer} per feature dimension, and Group Normalization \citep{wu2018group} over channel groups to handle settings where batch statistics are unreliable. These techniques operate on intermediate feature representations and are independent of the geometry of the input data.

The entity bias examined in this paper sits outside both of these frameworks. It arises in the spatial coordinate representation of skeleton sequences at the input level, prior to any feature extraction. It is not a cross-domain shift but a within-dataset asymmetry produced by a fixed convention for choosing a coordinate origin during data preprocessing. Standard feature normalization does not address this spatial asymmetry, and domain adaptation methods require explicit domain partitions that are unavailable in multi-entity interaction settings. CHASE is designed to close this gap. It learns a sample-adaptive spatial shift in the input space, constrained geometrically within the skeleton convex hull, and guided during training by a distribution alignment objective that operates over entities within each mini-batch.

\subsection{Skeleton-based Action Recognition}

\textbf{Models.} A significant body of work focuses on designing neural network architectures to improve skeleton-based action recognition. Early approaches primarily relied on recurrent architectures to capture long-term temporal contexts \citep{ST-LSTM2016, Co-LSTM2016, GCA2017, VA-LSTM2017, 2s-GCA2018, zhang2019view}. Subsequently, Graph Convolution Network (GCN)-based methods emerged as the dominant paradigm and remain central to this field today \citep{ST-GCN2018, AS-GCN2019, 2s-AGCN2019, MS-G3D2020, CTR-GCN2021, InfoGCN2022, hdgcn2023, 9329123, duan2022pyskl, liu2023tsgcnext, blockgcn2024CVPR, zhang2024shapmix, degcn2024tip, xie2025spatial}. To address specific challenges, various enhancements to GCNs have been proposed. InfoGCN \citep{InfoGCN2022} employs a novel objective to learn compact latent representations. BlockGCN \citep{blockgcn2024CVPR} proposes an efficient refinement to graph convolutions to mitigate redundancy issues. DeGCN \citep{degcn2024tip} introduces deformable sampling locations on spatiotemporal graphs, improving the perception of discriminative receptive fields. More recently, self-attention mechanisms have been incorporated into spatiotemporal modeling for skeletons, offering new perspectives on sequence encoding and interaction modeling \citep{dstanet2020, STSA-Net2023, zhou2022hypergraph, iiptrans2023, long2023step, MAMP_2023_ICCV, PSUMNet2023, do2024skateformer}. These approaches explore diverse tokenization strategies, attention designs, and training pipelines to enhance model performance. For instance, STSA-Net \citep{STSA-Net2023} employs a spatiotemporal segment encoding strategy to fuse joint relations across frames. Hyperformer \citep{zhou2022hypergraph} introduces a novel self-attention mechanism on hypergraphs, capturing higher-order relations intrinsic to skeleton data. To align with pretraining paradigms used in transformers, MAMP \citep{MAMP_2023_ICCV} performs masked self-component reconstruction on human joints, enabling effective feature representation for 3D action recognition.

\textbf{Representations.} Recent advancements in skeleton-based action recognition increasingly exploit various intra-skeleton modalities (i.e., skeletal representations) and their ensembles \citep{CTR-GCN2021,InfoGCN2022,hdgcn2023,dstanet2020,STSA-Net2023,degcn2024tip}, including joints, bones, and velocities. PSUMNet \citep{PSUMNet2023}, for instance, integrates multiple intra-skeleton modalities into a unified framework, demonstrating the effectiveness of such a holistic approach.

\textbf{Objectives.} There has also been interest in probing alternative or auxiliary optimization objectives to enhance robustness \citep{InfoGCN2022,huang2023SkeletonGCL}, incorporate supplementary textual descriptions \citep{xu2023language,xiang2023gap,he2024enhancing}, and address challenging scenarios \citep{Peng2024,liu2024recovering}.

However, these methods mainly focus on individual actions and circumvent multi-entity interaction problems. Taking late fusion strategy for adaptation relies on the assumption that entities are IID \citep{CTR-GCN2021,InfoGCN2022,hdgcn2023,dstanet2020,STSA-Net2023,2s-AGCN2019}. Our proposed approach can enhance these existing methods noticeably by addressing the entity bias problem.

\subsection{Skeleton-based Multi-Entity Action Recognition}

\textbf{Human-Human Interactions.} Since the introduction of two-person interaction datasets \citep{NTU60, NTU120, SBU}, the research community has recognized the importance of interaction modeling and has made significant efforts to develop more comprehensive benchmarks \citep{Guo_2022_CVPR, Yin_2023_CVPR, xu2024inter, Liang2024} and advanced models \citep{6890714, igformer2022, GDCN2023, LSTM-IRN2022, wen2023interactive, liu2024learning,chen2025ASEA}. For example, LSTM-IRN \citep{LSTM-IRN2022} incorporated relational reasoning to model interactions by capturing diverse relationships between human joints.
GDCN \citep{GDCN2023} introduced a graph diffusion convolutional network to uncover intrinsic local-global clues in two-person activities.
IGFormer \citep{igformer2022} leveraged prior knowledge of human body structure to design co-attention mechanisms for interaction recognition.
Similarly, me-GCN \citep{liu2024learning} extracted adjacency matrices for individual entities and modeled mutual constraints between them to enhance performance.
However, these approaches often rely on strong priors and intricate architectures that are specifically designed to handle interactions between exactly two human bodies, limiting their generalizability to broader scenarios.

\textbf{Hand-Object Interactions.} Many existing works has explored egocentric hand-hand \citep{Assembly101,Wen_2023_CVPR,Ohkawa_2023_CVPR,shamil2024HandFormer} and hand-object \citep{H2O_TA-GCN2021,Garcia-Hernando_2018_CVPR,H+O2019,shamil2024HandFormer,H2OTR2023CVPR,mucha2024perspective,zhu2025diagnosing} interactions. TA-GCN \citep{H2O_TA-GCN2021} employs a topology-aware graph convolutional network to model hand-object relationships, where the graph dependencies of hands are predefined as priors. EffHandEgoNet \citep{mucha2024perspective} introduces a transformer-based architecture for action recognition that utilizes 2D poses of hands and objects. Similarly, HandFormer \citep{shamil2024HandFormer} capitalizes on the unique characteristics of hand poses by temporally factorizing hand modeling and representing each joint through its short-term trajectories, achieving both efficiency and high accuracy. While these studies have made notable progress, they remain confined to this specific sub-domain by embedding hand and object priors directly into their model designs, which restricts their applicability to broader interaction scenarios.

\textbf{Human-Robot Interactions (HRI).} In recent years, the skeleton-based Human-Robot Interaction datasets have been introduced \citep{HARPER2024, chico2022}. HARPER \citep{HARPER2024} is the first dataset to capture physical dyadic interactions between humans and quadruped robots, including unexpected collisions. In contrast, Chico \citep{chico2022} focuses on industrial settings, providing a smaller-scale dataset with multi-view videos and skeletons of operators and cobots engaging in collaborative assembly tasks. Notably, these two datasets are the only available resources that offer both human and robot 3D skeletons along with sufficient interaction categories for model training. Despite their introduction, to the best of our knowledge, no prior work has specifically addressed the skeleton-based HRI recognition task, leaving a critical gap in this domain of research.

\textbf{Group Activities.} Group activities \citep{6095563, SkeleTR2023, WANG2024110478, che2024m3act, tamura2024design} involve assigning semantic labels to motions that often include a large number of entities, some of which may perform irrelevant individual actions \citep{cad2009, msibrahi2016VOL}. Methods tailored to this scenario \citep{Azar_2019_CVPR, Wu_2019_CVPR, Gavrilyuk_2020_CVPR, Li_2021_ICCV, Yuan_Ni_2021, tamura2022, Han_2022_CVPR, Thilakarathne2022-tm, zhou2022composer, 9710893} typically rely on multi-modality fusion or highly complex model architectures to achieve competitive performance. For example, COMPOSER \citep{zhou2022composer} employed extremely intricate multi-scale design alongside a series of specialized learning objectives to train a heavy model for only 10-frame group activity learning.

The above approaches excel in their respective sub-domains by leveraging two main strategies: incorporating domain-specific priors or employing intricate model designs. In contrast, we demonstrate that simple models, such as those initially designed for single-entity actions, can achieve competitive performance across these diverse scenarios when integrated with our CHASE method. Importantly, CHASE enables such models to perform effectively with minimal reliance on domain-specific priors, showcasing its adaptability.

\begin{figure*}[t]
    \begin{center}
    \includegraphics[width=\linewidth]{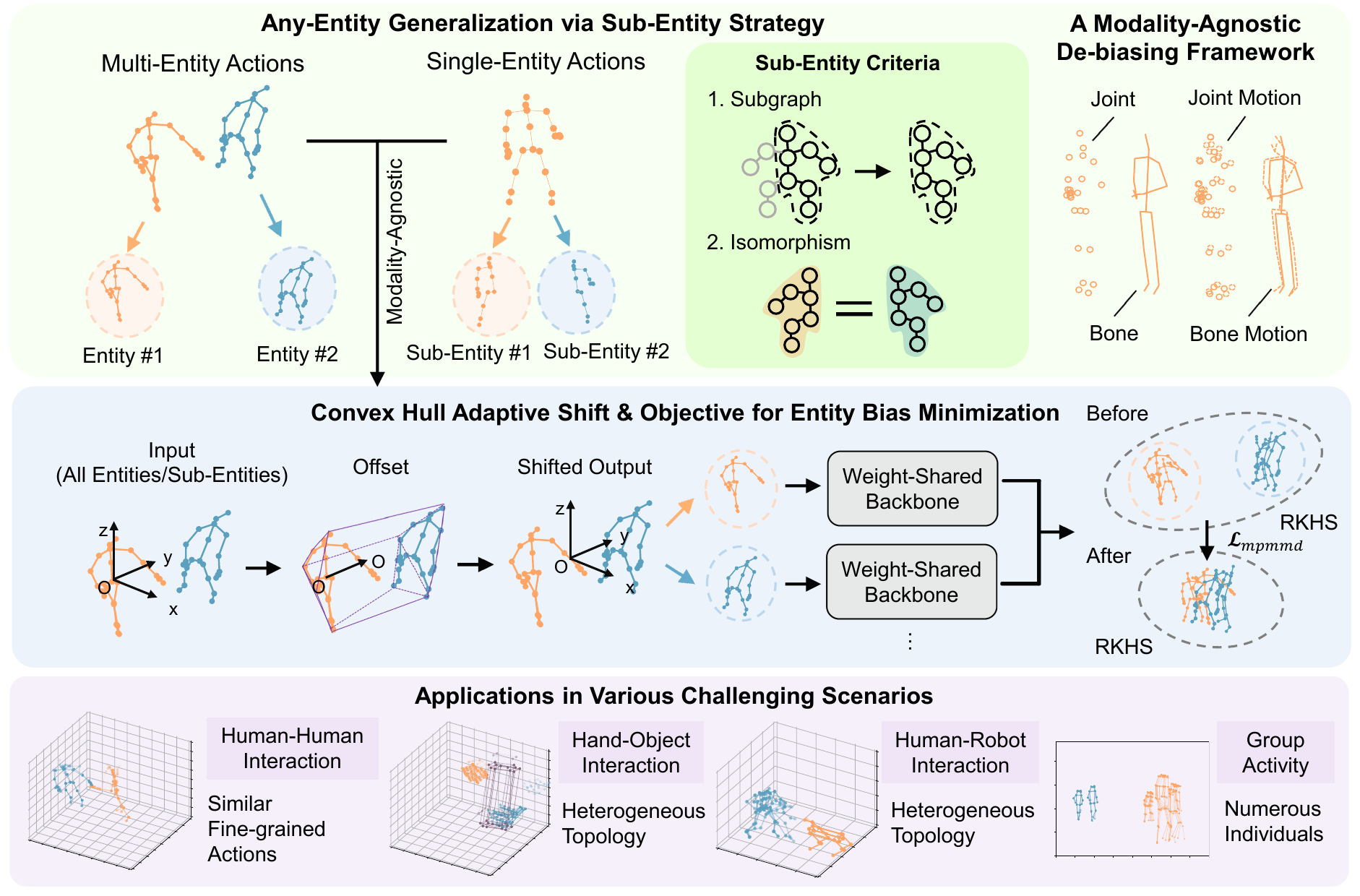}   
    \end{center}
    \caption{\textbf{The overall framework of the proposed CHASE for skeleton-based action recognition, including individual actions, person-person interactions, hand-object interactions, human-robot interactions, and group activities.} For clarity, we offer an example of person-person interaction skeleton sequence for pipeline illustration. Given a skeleton sequence as input, CHASE leverages Convex Hull Adaptive Shift, a learnable normalization module, to generate plausible and sample-adaptive offset for each input. It also collects pair-wise shifted skeletons within mini-batches, effectively addressing entity bias by introducing an auxiliary objective. Sub-Entity Strategy allows CHASE to be applied in both individual and multi-entity actions. Moreover, CHASE can be adopted in various widely-used skeletal modalities defined using joints and physical topology for de-biasing.}
    \label{fig:method}
    \vspace{-1.0em}
\end{figure*}

\section{CHASE}\label{sec:method}

Fig.~\ref{fig:method} presents the overall CHASE framework. The method is organized around four components, each building on the previous. Section~\ref{sec:Preliminaries} establishes the notation used throughout. Section~\ref{sec:ICHAS} presents the Convex Hull Adaptive Shift, the core normalization module, and Section~\ref{sec:mpmmd} introduces the auxiliary Entity Bias Minimization objective that guides its training. Together these two components form the foundation for multi-entity settings. The remaining two sections describe the journal-specific extensions. Section~\ref{sec:subentity} develops the sub-entity strategy, which reformulates the single-entity setting so that the same MPMMD objective remains applicable when only one entity is present. Section~\ref{sec:modality} examines how entity bias manifests across different skeletal representations and shows that CHASE operates without any modality-specific assumptions.

\subsection{Preliminaries}
\label{sec:Preliminaries}

\textbf{Skeleton Sequences.} Suppose that \(E\) entities (e.g. persons) engage in a purposeful act during a period of time \(T\), and the pose of each entity is indicated by \(J\) joints with \(C\) Cartesian coordinates. We can define the skeleton sequence of an action as \(X \in \mathbb{R}^{C\times T\times J \times E}\). For clarity, we denote the total number of points (across all frames, joints, and entities) as \(U=T\times J \times E\).

\textbf{Skeleton-based Action Recognition.} We define the task as finding the optimal estimator \(\mathcal{E}_{\theta}\) of the mapping \(\mathcal{E}:X\mapsto Y\), where \(X\) is the skeleton sequence of an action and \(Y\) is its corresponding label. 

\textbf{Late Fusion Strategy.} A common practice named \textit{late fusion strategy} is widely-adopted when single-entity action recognition models meet complicated skeletal data (e.g., with multiple entities) \citep{CTR-GCN2021,InfoGCN2022,hdgcn2023,dstanet2020,STSA-Net2023,degcn2024tip}. This strategy begins with separating each entity. Subsequently, weight-shared backbones (e.g., GCN and Transformer) learn spatiotemporal features of each entity. Finally, an averaging operator applies to these features, whose result serves as the overall feature of this skeleton sequence. Notably, this common practice uses weight-shared encoders, which implicitly relies on an empirical assumption that all entities are independent and identically distributed.

\subsection{Convex Hull Adaptive Shift}
\label{sec:ICHAS}

The bias observed in skeleton sequences arises from the initial configuration of the world coordinate system. A representative example is the standard preprocessing of NTU RGB+D 120 \citep{NTU120}, one of the most widely used benchmarks, in which the coordinate origin is translated to the spine-base joint (pelvis) of the first person. This convention is chosen for its stability as a reference point, and works well in single-entity recognition. In multi-entity settings, however, it introduces a systematic asymmetry: the first entity is always centered at the origin while all other entities are displaced by amounts proportional to their physical separation, producing inter-entity distributional discrepancies that scale with that separation. More broadly, any fixed choice of reference joint or reference entity during data collection or preprocessing will embed such asymmetries into the representation. To address this distribution discrepancy, we propose the Convex Hull Adaptive Shift, a learnable parameterized network characterized by two key features. First, the offset is adaptively tailored to each individual skeleton sequence. Second, the shift vector is inherently plausible, as it is implicitly constrained within the skeleton's convex hull. It avoids non-convergence by limiting the search space. By applying the Convex Hull Adaptive Shift, each sample is relocated to ensure that entities are rendered approximately IID.

For a skeleton sequence \(X\in \mathbb{R}^{C\times U}\), shifting all points by a common vector \(\Vec{p^*}\in \mathbb{R}^{C\times 1}\) can be written compactly as:
\begin{equation}
    \label{eq:sub_matrix}
    \hat{X} = X - \Vec{p^*}J_{1,U},
\end{equation}
where \(J_{1,U}\in \mathbb{R}^{1\times U}\) is a row vector of ones that broadcasts \(\Vec{p^*}\) across all \(U\) points, and \(\hat{X}\) is the shifted sequence. Making \(\Vec{p^*}\) depend on the input via \(\Vec{p^*}=XW\) introduces adaptability, but leaves \(\Vec{p^*}\) unconstrained in \(\mathbb{R}^{3}\), which can cause non-convergence. We address this by constraining \(\Vec{p^*}\) to lie within the convex hull of \(X\) \citep{Rockafellar1996} — the unique minimal convex set containing all points of \(X\), equivalently the set of all convex combinations of those points.

\begin{figure*}[t]
    \begin{center}
    \includegraphics[width=\linewidth]{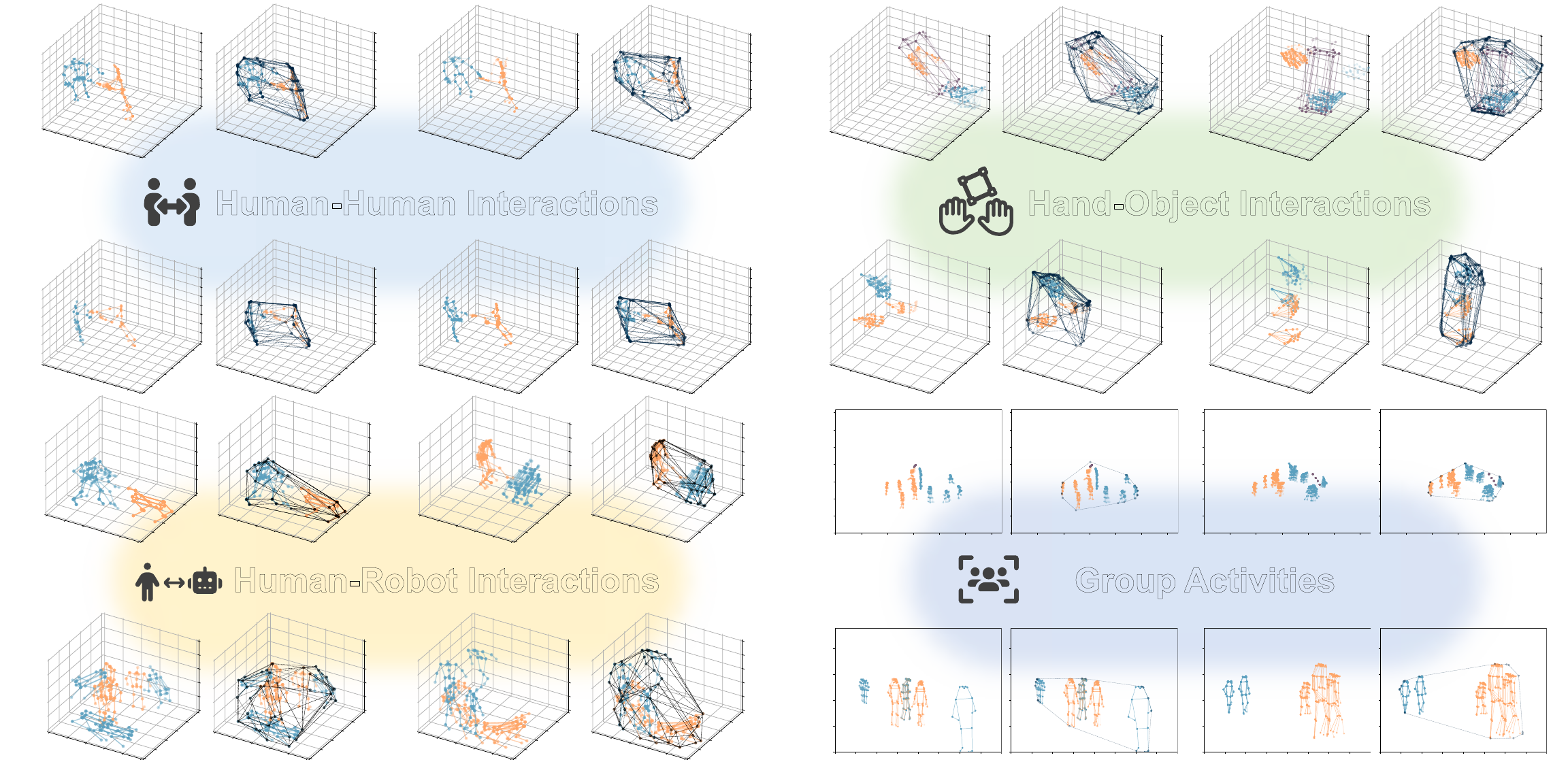}   
    \end{center}
    \vspace{-0.5em}
    \caption{\textbf{Skeleton convex hulls.} A skeleton convex hull, denoted using dark blue dashed lines, is defined by all the spatial and temporal keypoints in a skeleton sequence. All feasible new origins remain within the convex hull, which limits the search space for shift vector and avoids non-convergence.}
    \label{fig:dataset}
    \vspace{-1.0em}
\end{figure*}

The shift vector \(\Vec{p^*}\) is formulated as:
\begin{equation}
    \label{eq:xsoftmaxw}
    \Vec{p^*} = X\mathrm{softmax}(W),
\end{equation}
where \(X\in \mathbb{R}^{C\times U}\), \(W\in \mathbb{R}^{U\times 1}\), and \(\Vec{p^*}\in \mathbb{R}^{C\times 1}\). Let \(\tilde{\alpha}_i = e^{\alpha_i}/\sum_{j}e^{\alpha_j}\) denote the \(i\)-th component of \(\mathrm{softmax}(W)\), where \(\alpha_i\) is the \(i\)-th element of \(W\). Since \(\tilde{\alpha}_i\in(0,1)\) and \(\sum_i\tilde{\alpha}_i=1\), we have \(\Vec{p^*}=\sum_{i=1}^{U}\tilde{\alpha}_i\Vec{p}_i\), which satisfies the definition of a convex combination. Therefore, \(\Vec{p^*}\) represents a convex combination of points in \(X\), lying within the minimal convex set encompasses \(X\), as shown in Fig.~\ref{fig:dataset}.

\textbf{Convex Hull Adaptive Shift (CHAS).} Combining Formula (\ref{eq:sub_matrix}) and (\ref{eq:xsoftmaxw}), we propose the Convex Hull Adaptive Shift as:
\begin{equation}
    \label{eq:ConAdaShf}
    \hat{X} = X(I - \mathrm{softmax}(W)J_{1,U}),
\end{equation}
where \(I\in \mathbb{R}^{U\times U}\) is the identity matrix. This formulation restricts the search space of \(\Vec{p^*}\) from the entire \(\mathbb{R}^{3}\) to the open convex hull, ensuring both adaptivity and plausibility.

\textbf{Coefficient Learning Block (CLB).} With a fixed \(W\), the gradient \(\partial\hat{X}/\partial X = I - J_{U,1}\mathrm{softmax}(W^T)\) is constant, meaning all inputs share the same shift coefficients. True sample-adaptivity therefore requires \(W\) to depend on \(X\) through a learnable mapping \(\psi:\mathbb{R}^{C\times U}\mapsto \mathbb{R}^{U\times 1}\). As illustrated in Fig.~\ref{fig:method}, we implement this as a lightweight Coefficient Learning Block:
\begin{equation}
    \label{eq:psi}
    W=\psi(X)=W_3\delta(W_2\phi(W_1X+b)),
\end{equation}
where \(W_1\in \mathbb{R}^{C_1\times C}, W_2\in \mathbb{R}^{C_2\times C_1}, W_3\in \mathbb{R}^{U\times C_2}\) are weight matrices, \(b\) is a bias, \(\phi\) is a squeeze operator \citep{hu2019senet}, and \(\delta\) is an activation function. We set \(U\geq C_1>C_2\) following the bottleneck design \citep{hu2019senet,NIPS2015_215a71a1}.

\subsection{Objective for Entity Bias Minimization}
\label{sec:mpmmd}
To optimize CHAS, we introduce an auxiliary objective that minimizes pairwise distributional discrepancies between entities, thereby preventing the learned shifts from collapsing into trivial solutions. Enforcing alignment between the feature distributions of different entities within each training mini-batch incrementally drives the network toward a canonical, shared coordinate space — mitigating entity bias without requiring global distribution matching.

We build this objective on Maximum Mean Discrepancy (MMD), a kernel-based metric that measures the distance between two distributions \(P(x)\) and \(Q(y)\) as:
\begin{equation}
    \label{eq:vanimmd}
    \text{MMD}(P,Q)=\sup _{\|f\|_{\mathcal {H}}\leq 1}\left(\mathbb {E} [f(x)]-\mathbb {E} [f(y)]\right),
\end{equation}
where the supremum is over all functions \(f\) in the reproducing kernel Hilbert space \(\mathcal{H}\). For \(E\) entity distributions \(P^i\), the average pairwise MMD is:
\begin{equation}
\label{eq:mmd}
    \mathbb{E}_{r(z)}[\text{MMD}(z)]=\sum_{i=1}^{E-1}\sum_{j=i+1}^{E}\text{MMD}(P^i,P^j)/\text{C}(E,2),
\end{equation}
where \(z=(P^i,P^j)\), \(r(z)\) is the probability density over pairs, and \(C(E,2)\) counts all entity pairs. Two approximations make this tractable: expectations are estimated over mini-batches, and rather than enumerating all \(C(E,2)\) pairs (complexity \(O(n!)\)), we sample \(M\) pairs uniformly, giving:
\begin{equation}
    \label{eq:MPMMD}
    \mathbb{E}_{r(z)}[\text{MMD}(z)] \approx \frac{1}{M}\sum_{m=1}^{M}\text{MMD}(z_m).
\end{equation}

We refer to this as the Mini-batch Pairwise Maximum Mean Discrepancy Loss \(\mathcal{L}_{mpmmd}\). The total training loss is:
\begin{equation}
    \label{eq:totalloss}
    \mathcal{L} = \mathcal{L}_{CLS} + \lambda \mathcal{L}_{mpmmd},
\end{equation}
where \(\lambda\) is a trade-off factor and \(\mathcal{L}_{CLS}\) is for classification.

The pseudo-code implementation of CHASE is detailed in Algorithm~\ref{algo:chaswrapper}. For the current mini-batch, we apply a learnable normalization module to the input skeletal data \(x\), and guide the bias minimization with the proposed MPMMD loss.

\begin{algorithm}[t]
\caption{Pseudo code of CHASE in a PyTorch-like style.}
\label{algo:chaswrapper}
\algcomment{\fontsize{7.2pt}{0em}\selectfont \texttt{prod}: Product of all elements; \texttt{hd}: Hadamard Division; \texttt{ce}: Cross-Entropy.
}
\definecolor{codeblue}{rgb}{0.25,0.5,0.5}
\lstset{
  backgroundcolor=\color{white},
  basicstyle=\fontsize{7.2pt}{7.2pt}\ttfamily\selectfont,
  columns=fullflexible,
  breaklines=true,
  captionpos=b,
  commentstyle=\fontsize{7.2pt}{7.2pt}\color{codeblue},
  keywordstyle=\fontsize{7.2pt}{7.2pt},
}
\begin{lstlisting}[language=python]
# backbone: encoder networks for single entity
# in_channels: number of coordinates (C) 
# num_frame: number of frames (T)
# num_point: number of joints per entity (J)
# num_entity: number of entities (E) 
# pooling_seg, c1, c2: projection dimensions
# lambda: trade-off factor

# initialize
num_list = [num_frame, num_point, num_entity]
out_channel = prod(num_list)
seg = prod(pooling_seg)
seg_num_list = round(hd(num_list, pooling_seg))
seg_num = prod(seg_num_list)
shift = nn.Sequential(
    nn.Conv3d(in_channels, c1, 1),
    nn.AdaptiveAvgPool3d(pooling_seg),
    nn.Conv3d(c1, c2, 1, bias=False),
    nn.ReLU(inplace=True),
    nn.Conv3d(c2, outchannel, 1, bias=False),
)

for x, label in loader:  # load a minibatch
    N, C, T, V, M = x.size()
    # Formula (7)
    sf = shift(x).view(N, T*M*V, -1)
    tx = rearrange(x)
    # Formula (5)
    sf = (tx @ sf.softmax(dim=1)).unsqueeze(-1)
    sf = rearrange(sf)
    x = x - sf
    # Formula (10)
    pairs = MiniBatchSampling(x)
    aux_loss = MPMMD(pairs)
    out = backbone(x)
    # Formula (11)
    total_loss = ce(out, label)+lambda*aux_loss
\end{lstlisting}
\end{algorithm}

\subsection{Any-Entity Generalization via Sub-Entity Strategy}
\label{sec:subentity}

\begin{figure}[t]
    \begin{center}
    \includegraphics[width=\linewidth]{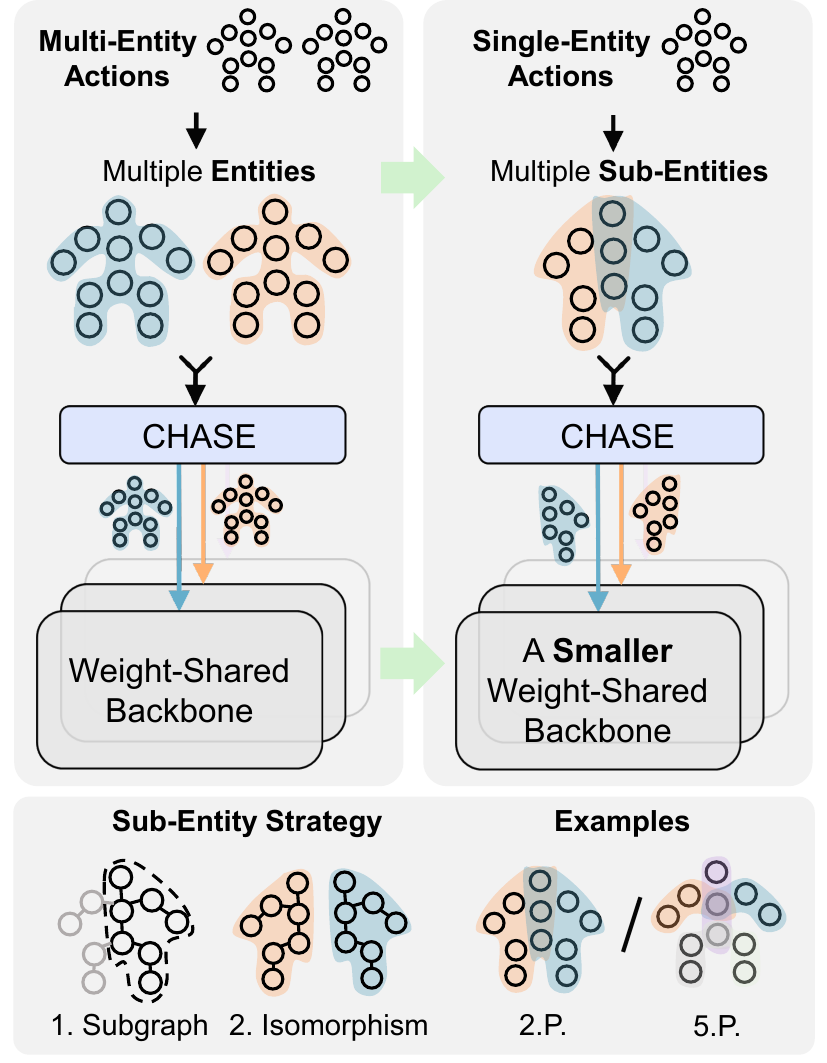}   
    \end{center}
    \vspace{-0.5em}
    \caption{\textbf{Overview of sub-entity strategy}. Black circles denote joints. A single entity can be viewed as many sub-entities of its own parts (denoted in different colors), which satisfy the subgraph and isomorphism conditions. Examples on human body skeletons are 2-part (2.P.) and 5-part (5.P.) partitions.}
    \label{fig:subentity}
    \vspace{-1.5em}
\end{figure}

So far, we have discussed CHASE under the condition \(E\geq 2\). But what happens when \(E=1\)? Single-entity actions are a critical area in spatiotemporal motion understanding. However, without multiple entities, defining pair-wise distribution discrepancies becomes infeasible. To address this, we introduce a sub-entity strategy that extends CHASE to single-entity actions. This approach adapts multi-entity definitions, providing a unified framework for handling any-entity actions.

\textbf{Sub-Entity Strategy.} In this context, the CHASE framework treats a single entity as composed of multiple sub-entities. Accordingly, in the formulations that follow (Equations \eqref{eq:ConAdaShf} and related), \(U\) denotes the number of these sub-entities, rather than the total number of temporal and spatial points. The sub-entity strategy is equivalent to selecting sub-entities w.r.t. a given graph \(G=(V,E)\), such that the chosen sub-entities satisfy the following conditions: (1) Each sub-entity is a subgraph of the original graph; (2) All sub-entities are isomorphic. These conditions are detailed as follows:

1) \textit{Subgraph}: 
Let \( G = (V, E) \) be the original graph, and \( G' = (V', E') \) be a sub-entity. Then, \( G' \) is a subgraph of \( G \) if and only if:
\begin{equation}
   V' \subseteq V , E' \subseteq E, \quad \text{and} \quad \forall (u, v) \in E', u, v \in V'.
\end{equation}
This means that \( G' \) consists of a subset of the nodes \( V' \) and a subset of the edges \( E' \), where every edge in \( E' \) connects nodes in \( V' \).

2) \textit{Isomorphism}: 
Let \( G_1 = (V_1, E_1) \) and \( G_2 = (V_2, E_2) \) be two sub-entities. Then, \( G_1 \) and \( G_2 \) are isomorphic if there exists a bijection \( g: V_1 \to V_2 \) such that for all pairs of nodes \( u, v \in V_1 \), the following holds:
\begin{equation}
   (u, v) \in E_1 \iff (g(u), g(v)) \in E_2.
\end{equation}
This means there is a one-to-one correspondence \( g \) between the node sets of \( G_1 \) and \( G_2 \), such that the adjacency relationships (edges) are preserved under this mapping.

Sub-entities are derived from physical topology. Figure~\ref{fig:subentity} shows two examples: 2-part (2.P.) and 5-part (5.P.) decompositions based on natural body structure \citep{xiang2023gap}. Sub-entities may overlap and exhibit 3D chirality; they can be proper or improper subgraphs.

The primary motivation of this strategy differs fundamentally from conventional part-based modeling in skeleton-based recognition. Part-based methods decompose the body to improve feature extraction or capture local motion patterns in the backbone. The sub-entity strategy, by contrast, is not concerned with how features are extracted. Its purpose is to construct a set of structurally comparable units from a single entity so that the MPMMD loss can compute meaningful pairwise distribution discrepancies, an operation that is otherwise undefined when only one entity is present. The isomorphism condition is the critical formal requirement that makes this possible: it guarantees that all sub-entities share the same graph structure, and are therefore exchangeable under the distribution matching objective. Without this constraint, comparing distributions across structurally heterogeneous parts would be ill-defined. For single-entity actions (e.g., jumping), we treat an entity as multiple sub-entities corresponding to its body parts. This also enables smaller, weight-shared backbones by reducing the joint count per sub-entity, improving both efficiency and performance.

The mechanism by which CHASE reduces entity bias among body parts follows the same principle as in the multi-entity setting. Even within a single person, anchoring the coordinate origin to a fixed reference point introduces distributional asymmetries between different body regions: parts that are spatially far from the reference accumulate larger positional offsets than parts that are close to it, so their joint coordinate distributions diverge in location and spread across samples. CHASE learns a sample-adaptive shift that brings these intra-body distributions closer together, guided by the MPMMD objective operating on the isomorphic sub-entities. The intra-body asymmetry is smaller in magnitude than the inter-person asymmetry, because body parts of the same person are spatially closer to each other than different persons typically are, which explains why the performance gains in single-entity settings are consistent but more modest than those in multi-entity settings.

\subsection{A Modality-Agnostic De-biasing Framework}
\label{sec:modality}

The above sections formulate CHASE for joint modality (physical joints or keypoints). However, skeleton-based action recognition increasingly uses multiple intra-skeleton modalities and their ensembles \citep{CTR-GCN2021,InfoGCN2022,hdgcn2023,dstanet2020,STSA-Net2023,degcn2024tip}, such as bones and velocities. Here, we extend CHASE to these modalities.

The definitions of the commonly used skeletal modalities (i.e., representations) are listed below:

\begin{itemize}
    \item Joint: A joint in a skeleton sequence is a point in the Cartesian coordinate system, which can be viewed as a vector \(\Vec{p}_{i,t} \in \mathbb{R}^{C\times 1}\), where \(1\leq i\leq J\) and \(1\leq t \leq T\).
    \item Bone: A bone in a skeleton sequence is a differential vector \(\Vec{p}_{i,j,t}=\Vec{p}_{i,t}-\Vec{p}_{j,t}\), connecting two joints \(\Vec{p}_{i,t}\) and \(\Vec{p}_{j,t}\). Here, \(1 \leq i, j \leq J\), \(1 \leq t \leq T\), and the connection between the joints is determined by a predefined graph (e.g., the bone structure of the human body). 
    \item Velocity: A velocity vector is a differential vector \(\Vec{p}_{i,t,s}=\Vec{p}_{i,t}-\Vec{p}_{i,s}\), representing the movement of the same joint (or bone) across different timestamps \(t\) and \(s\), where typically \(t = s + 1\) and \(1 \leq s \leq T - 1\).
    \item JBF \citep{degcn2024tip}: A JBF vector is defined as the concatenation of a joint and a bone, represented as \(\Vec{p}_{jbf} \in \mathbb{R}^{2C \times 1}\).  
\end{itemize}

Since vectors are points, the above definitions yield a set of points \(X'\in \mathbb{R}^{C'\times U'}\), where dimensions depend on the skeletal modality. CHASE only requires point sets as input, enabling straightforward extension to any modality. We reformulate Formula (\ref{eq:ConAdaShf}) as:
\begin{equation}
    \hat{X} = X'(I - \mathrm{softmax}(W)J_{1,U'}).
\end{equation}
Formulas (\ref{eq:psi}) and (\ref{eq:MPMMD}) remain valid by replacing \(X\) with \(X'\). For bone representations, the sub-entity strategy preserves the Subgraph and Isomorphism conditions using the line graph \(L(G)\), the edge-to-vertex dual of \(G\). In \(L(G)\), each edge of \(G\) becomes a vertex, and edges connect vertices representing edges in \(G\) that share a common vertex. Sub-entities are then selected as subgraphs of \(L(G)\).

\begin{table*}[t]
	\centering
	\caption{Details of Benchmarks Used in Experiments}
	\label{table:dataset}
        \resizebox{\textwidth}{!}{
        \begin{threeparttable}
	\begin{tabular}{l|c|c|c|c|c|c|c|c|c|c}
		\hline
            \multirow{2}{*}{Benchmark}&\multicolumn{4}{c|}{Skeleton Sequences}&\multirow{2}{*}{\#Category}&\multirow{2}{*}{\#Joint}&\multirow{2}{*}{\#Entity}&\multirow{2}{*}{\#Participant}&\multirow{2}{*}{Type}&\multirow{2}{*}{Example(s)}\\	
            \cline{2-5}
            &Body&Hand&Object&Robot&&&&&&\\
		\hline
            NTU60 \citep{NTU60}&\Checkmark&&&&60&25(\(\times\)2)&1 or 2&40&3D&\textit{hugging}, \textit{walking towards}\\
            NTU120 \citep{NTU120}&\Checkmark&&&&120&25(\(\times\)2)&1 or 2&106&3D&\textit{exchange things}, \textit{whisper}\\	
            H2O \citep{H2O_TA-GCN2021}&&\Checkmark&\Checkmark&&36&21\(\times\)3&3&4&3D&\textit{squeeze lotion out from bottle}\\
            Assembly101 \citep{Assembly101}&&\Checkmark&&&1,380&21\(\times\)2&2&53&3D&\textit{screw track with hand}\\	
            CAD \citep{cad2009}&\Checkmark&&&&4&17\(\times\)5.22&\(\approx\)5.22&Uncontrolled&2D&\textit{queuing}, \textit{walking}, \textit{talking}\\
            VD \citep{msibrahi2016VOL}&\Checkmark&&\Checkmark&&8&17\(\times\)12+1&13&Uncontrolled&2D&\textit{left pass}, \textit{right spike}\\
            HARPER \citep{HARPER2024}&\Checkmark&&&\Checkmark&15&21+23&2&17&3D&\textit{circular follow + crash}\\
            \hline
	\end{tabular}
       \end{threeparttable}
        }
\end{table*}

\section{Experiments}\label{sec:exp}

\subsection{Datasets \& Evaluation}

In experiments, we evaluate the methods on a variety of skeletal datasets, including human actions \& interactions \citep{NTU60,NTU120}, hand-object interactions \citep{H2O_TA-GCN2021}, bi-hand interactions \citep{Assembly101}, human-robot interactions \citep{HARPER2024}, and group activities \citep{cad2009,msibrahi2016VOL}. Details of these datasets are listed in Table~\ref{table:dataset}.

\textbf{NTU RGB+D (NTU60)} \citep{NTU60} is a widely-used dataset for human activity recognition, structured into two distinct subsets: \textbf{NTU49}, which focuses on actions involving a single individual, and \textbf{NTU11}, dedicated to interactions between two people. In our experiments, we adopt the commonly applied X-Sub and X-View evaluation protocols.

\begin{table*}[t]
    \renewcommand\arraystretch{1.2}
	\centering
	\caption{Comparisons with Related Methods on Multi-Entity Action Recognition Datasets. Our proposed CHASE can improve diverse single-entity backbones to achieve state-of-the-art performance across a variety of benchmarks.}
	\label{tab:sota}
        \resizebox{\textwidth}{!}{
        \begin{threeparttable}
        \resizebox{\textwidth}{!}{
	\begin{NiceTabular}{l|c|C{10mm}|C{10mm}|C{10mm}|C{12mm}|c|c|c|c|c}[colortbl-like]
		\hline
		  \multicolumn{1}{l}{\multirow{2}{*}{Method*}}&
            \multirow{2}{*}{Venue}&
            \multicolumn{2}{c|}{NTU26(\%)}&
            \multicolumn{2}{c|}{NTU11(\%)}&
            \multirow{2}{*}{H2O(\%)}&
            \multirow{2}{*}{ASB101(\%)}&
            \multirow{2}{*}{CAD(\%)}&
            \multirow{2}{*}{VD(\%)}&
            \multirow{2}{*}{HARPER(\%)}\\
            \cline{3-6}

		  &
            &
            X-Sub&
            X-Set&
            X-Sub&
            X-View&
            &
            &
            &
            &
            \\

            \hline

            \multicolumn{12}{c}{Multi-Entity Backbones}
            \\

            \hline

            \rowcolor{myYellow!15}
            LSTM-IRN \citep{LSTM-IRN2022}&
            TMM'22&
            77.70&
            79.60&
            90.50&
            93.50&
            -&
            -&
            -&
            -&
            -\\

            \rowcolor{myYellow!15}
            IGFormer \citep{igformer2022}&
            ECCV'22&
            85.40&
            86.50&
            93.60&
            96.50&
            -&
            -&
            -&
            -&
            -\\
 
            \rowcolor{myYellow!15}
            SkeleTR \citep{SkeleTR2023}&
            ICCV'23&
            87.80&
            88.30&
            -&
            -&
            -&
            -&
            -&
            -&
            -\\

            \rowcolor{myYellow!15}
            ISTA-Net \citep{wen2023interactive}&
            IROS'23&
            90.56&
            91.72&
            -&
            -&
            89.09&
            28.01&
            87.16&
            91.40&
            79.63\\

            \rowcolor{myYellow!15}
            H2OTR \citep{H2OTR2023CVPR}&
            CVPR'23&
            -&
            -&
            -&
            -&
            90.90&
            -&
            -&
            -&
            -\\

            \rowcolor{myYellow!15}
            GDCN \citep{GDCN2023}&
            TPAMI'23&
            85.80&
            92.10&
            -&
            -&
            -&
            -&
            -&
            -&
            -\\  

            \rowcolor{myYellow!15}
            AHNet-Large \citep{WANG2024110478}&
            PR'24&
            86.43&
            86.64&
            90.85&
            93.38&
            -&
            -&
            89.32&
            84.31&
            -\\

            \rowcolor{myYellow!15}
            ME-Former \citep{meformer2024}&
            Biomi.'24&
            90.84&
            91.33&
            95.37&
            97.60&
            -&
            -&
            -&
            -&
            -\\

            \rowcolor{myYellow!15}
            EffHandEgoNet \citep{mucha2024perspective}&
            arXiv'24&
            -&
            -&
            -&
            -&
            91.32&
            -&
            -&
            -&
            -\\

            \rowcolor{myYellow!15}
            HandFormer \citep{shamil2024HandFormer}&
            ECCV'24&
            -&
            -&
            -&
            -&
            57.44&
            28.80&
            -&
            -&
            -\\

            \rowcolor{myYellow!15}
            me-GCN \citep{liu2024learning}&
            THMS'25&
            90.00&
            90.00&
            95.50&
            98.20&
            -&
            27.20&
            -&
            -&
            -\\

            \rowcolor{myYellow!15}
            ASEA \citep{chen2025ASEA}&
            ACMMM'25&
            90.52&
            91.77&
            -&
            -&
            -&
            -&
            -&
            -&
            -\\

            \hline

            \multicolumn{12}{c}{Single-Entity Backbones (+ CHASE)}
            \\

            \hline

            \rowcolor{myBlue!10}
            CTR-GCN \citep{CTR-GCN2021}&
            ICCV'21&
            89.32&
            90.19&
            95.94&
            98.32&
            81.68&
            27.83&
            80.45&
            92.66&
            78.24\\

            \rowcolor{myBlue!20}
            \textbf{+ CHASE (Ours)}&
            -&
            {\color{myHighlight}91.30}&
            {\color{myHighlight}92.34}&
            {\color{myHighlight}96.45}&
            \multicolumn{1}{c}{\textbf{{\color{myHighlight}98.83}}}&
            {\color{myHighlight}91.05}&
            {\color{myHighlight}28.03}&
            \multicolumn{1}{c}{\textbf{{\color{myHighlight}89.61}}}&
            {\color{myHighlight}92.89}&
            {\color{myHighlight}86.11}\\

            \rowcolor{myBlue!10}
            InfoGCN \citep{InfoGCN2022}&
            CVPR'22&
            90.22&
            91.13&
            95.51&
            97.76&
            76.24&
            27.18&
            83.07&
            91.77&
            80.56\\

            \rowcolor{myBlue!20}
            \textbf{+ CHASE (Ours)}&
            -&
            \multicolumn{1}{c}{\textbf{{\color{myHighlight}91.86}}}&
            {\color{myHighlight}92.41}&
            {\color{myHighlight}96.35}&
            {\color{myHighlight}98.25}&
            {\color{myHighlight}83.47}&
            {\color{myHighlight}27.36}&
            {\color{myHighlight}84.18}&
            {\color{myHighlight}92.00}&
            {\color{myHighlight}83.33}\\

            \rowcolor{myBlue!10}
            STSA-Net \citep{STSA-Net2023}&
            Neuro.'23&
            88.41&
            90.19&
            95.96&
            98.47&
            92.29&
            27.70&
            80.20&
            92.52&
            78.71\\

            \rowcolor{myBlue!20}
            \textbf{+ CHASE (Ours)}&
            -&
            {\color{myHighlight}89.77}&
            {\color{myHighlight}91.54}&
            {\color{myHighlight}96.63}&
            {\color{myHighlight}98.73}&
            \multicolumn{1}{c}{\textbf{{\color{myHighlight}94.77}}}&
            {\color{myHighlight}27.81}&
            {\color{myHighlight}85.93}&
            {\color{myHighlight}92.78}&
            {\color{myHighlight}88.89}\\

            \rowcolor{myBlue!10}
            HD-GCN \citep{hdgcn2023}&
            ICCV'23&
            88.25&
            90.08&
            95.58&
            97.93&
            72.73&
            27.31&
            76.93&
            91.32&
            80.10\\

            \rowcolor{myBlue!20}
            \textbf{+ CHASE (Ours)}&
            -&
            {\color{myHighlight}90.81}&
            {\color{myHighlight}92.06}&
            {\color{myHighlight}96.22}&
            {\color{myHighlight}98.31}&
            {\color{myHighlight}81.61}&
            {\color{myHighlight}27.50}&
            {\color{myHighlight}82.39}&
            {\color{myHighlight}92.00}&
            {\color{myHighlight}89.82}\\

            \rowcolor{myBlue!10}
            DeGCN \citep{degcn2024tip}&
            TIP'24&
            90.84&
            90.36&
            95.94&
            97.95&
            80.16&
            28.76&
            83.66&
            93.74&
            83.79\\

            \rowcolor{myBlue!20}
            \textbf{+ CHASE (Ours)}&
            -&
            {\color{myHighlight}91.61}&
            \multicolumn{1}{c}{\textbf{{\color{myHighlight}92.77}}}&
            \multicolumn{1}{c}{\textbf{{\color{myHighlight}96.82}}}&
            {\color{myHighlight}98.75}&
            {\color{myHighlight}86.50}&
            \multicolumn{1}{c}{\textbf{{\color{myHighlight}28.91}}}&
            {\color{myHighlight}86.41}&
            \multicolumn{1}{c}{\textbf{{\color{myHighlight}93.97}}}&
            \multicolumn{1}{c}{\textbf{{\color{myHighlight}94.90}}}\\

            \rowcolor{myBlue!10}
            Hyper-GCN \citep{zhou2025adaptive}&
            ICCV'25&
            86.90&
            88.46&
            94.97&
            97.67&
            80.44&
            19.35&
            65.51&
            79.40&
            72.69\\

            \rowcolor{myBlue!20}
            \textbf{+ CHASE (Ours)}&
            -&
            {\color{myHighlight}89.01}&
            {\color{myHighlight}90.29}&
            {\color{myHighlight}96.12}&
            {\color{myHighlight}98.68}&
            {\color{myHighlight}90.22}&
            {\color{myHighlight}21.04}&
            {\color{myHighlight}70.74}&
            {\color{myHighlight}81.41}&
            {\color{myHighlight}84.26}\\
		
		\hline
	\end{NiceTabular}
        }
        \begin{tablenotes}
         \item * The best result of each benchmark is \textbf{{\color{myHighlight}in bold}}.
        \end{tablenotes}
        \end{threeparttable}
        }
\end{table*}

\textbf{NTU RGB+D 120 (NTU120)} \citep{NTU120}, an expanded version of NTU60, offers a larger and more diverse collection of human activity samples. It is also divided into two non-overlapping subsets: \textbf{NTU94} for individual actions and \textbf{NTU26} for mutual activities involving two participants. For evaluation, we utilize the X-Sub and X-Set benchmarks as described in \citep{NTU120}.

\textbf{H2O} \citep{H2O_TA-GCN2021} contributes to the field of 3D vision by offering annotated human hand poses and bounding boxes for manipulated objects. These resources support the modeling of both hand-object and hand-hand interactions. We utilize the training, validation, and test divisions as defined in \citep{H2O_TA-GCN2021}.

\textbf{Assembly101 (ASB101)} \citep{Assembly101}, a comprehensive dataset for 3D manual procedural activities, features 1,380 distinct categories of interactive actions. Following \citep{Assembly101}, we adopt its predefined splits for training, validation, and testing. The labels used in our evaluations include detailed verb and noun pairs to capture fine-grained action semantics.

\textbf{Collective Activity Dataset (CAD)} \citep{cad2009} documents pedestrian activities in public spaces using camera footage. It organizes group behaviors into four categories and assigns individual action labels. Our implementation follows the same classification schema and train-test configuration described in \citep{zhou2022composer}, but relying solely on 2D joint coordinate data.

\textbf{Volleyball Dataset (VD)} \citep{msibrahi2016VOL} contains match footage from volleyball games, where group activities are categorized into eight classes aligned with volleyball terminology. For our experiments, we adhere to the original splits proposed in \citep{zhou2022composer}, but relying solely on 2D joint coordinate data.

\textbf{HARPER} \citep{HARPER2024} is a pioneering dataset centered on human-robot interaction, showcasing 15 types of dyadic interactions between humans and a Boston Dynamics Spot quadruped robot. Performed by 17 individuals, these interactions predominantly involve physical contact. We use the training and testing splits recommended in \citep{HARPER2024} for our evaluations.

\begin{figure*}[t]
    \begin{center}
    \includegraphics[width=\linewidth]{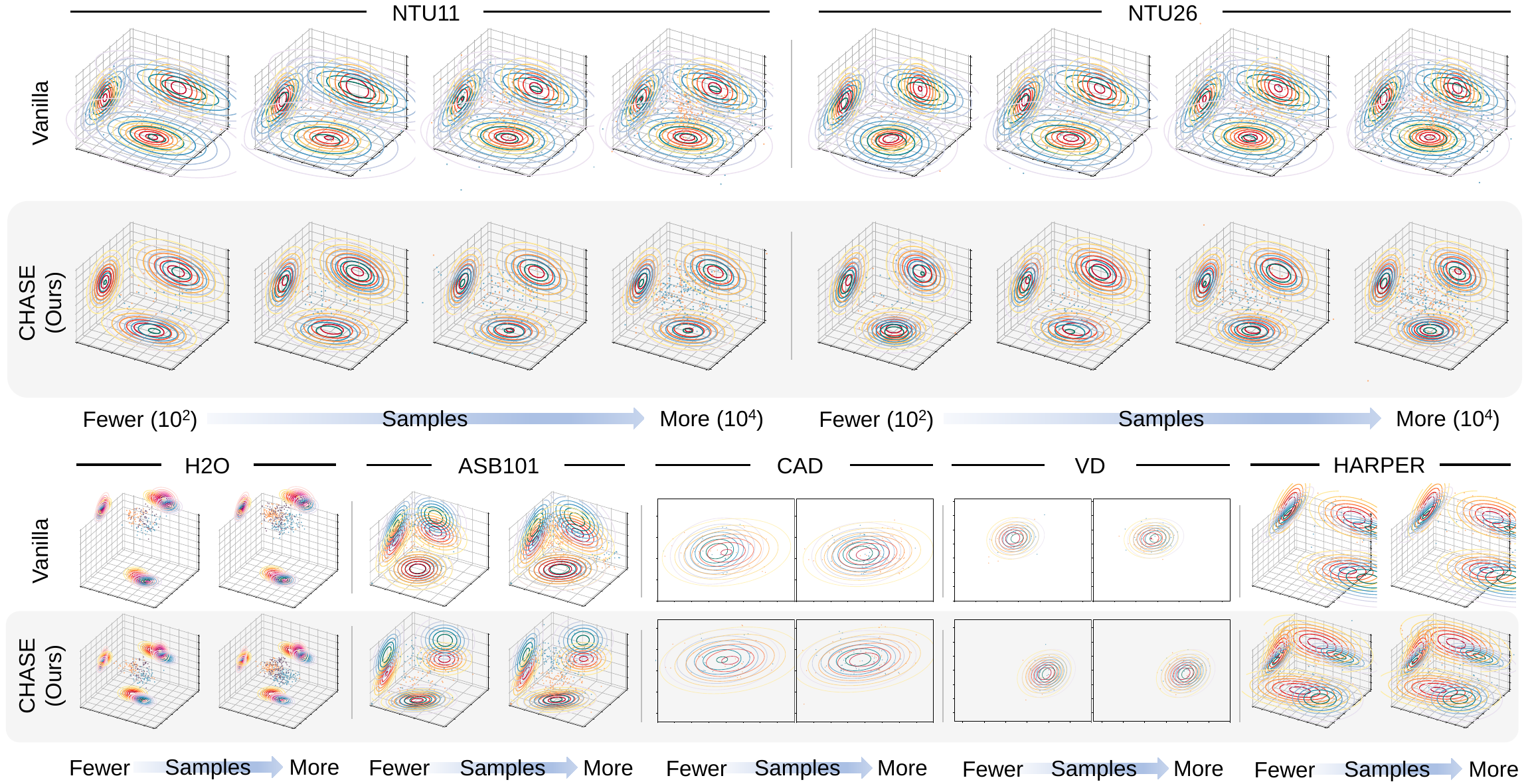}
    \end{center}
    \caption{\textbf{Qualitative results of CHASE}. Different entity distributions are denoted by blue and orange. By incorporating CHASE, entity distributions become more aligned in terms of both mean and covariance. CHASE effectively mitigates the inherent entity bias, demonstrating its clear effectiveness across a range of data scales and a variety of benchmarks.}
    \label{fig:vizchas}
\end{figure*}

\subsection{Implementation Details}

Experiments are conducted on the GeForce RTX 3070 GPUs. We select a diverse set of baseline models, encompassing both GCN-based, hyper-GCN-based, and transformer-based architectures. Specifically, our GCN-based baselines include CTR-GCN \citep{CTR-GCN2021}, InfoGCN \citep{InfoGCN2022} (k=1), HD-GCN \citep{hdgcn2023} (CoM=1), and DeGCN \citep{degcn2024tip}. Hyper-GCN (large) \citep{zhou2025adaptive} is chosen as the hyper-GCN-based baseline. STSA-Net \citep{STSA-Net2023} is included as our transformer-based representative. For training on NTU26 dataset, we follow data pre-processing and augmentations adopted in \citep{CTR-GCN2021,wen2023interactive}. The classification loss is implemented by the cross entropy with label smoothing factor 0.1. SGD optimizer is used with Nesterov momentum of 0.9, a initial learning rate of 0.1 and a decay rate 0.1. Batch size is 64. Each training process was terminated after 110 epochs. As the hyper-parameters may vary across different benchmarks, please refer to the benchmark-specific configurations in our code repository.

\begin{table}[t]
	\centering
	\caption{Comparison with Other Alternatives}
	\label{ablation:alternatives}
	\begin{tabular}{l|c|c}
            \hline
                Method (w/ CTR-GCN) & Acc (\%) & \(\Delta\) (\%) \\
		\hline
                Vanilla & 89.32 & - \\
            \hline
                S2CoM & 88.66 & -0.67\\
                BatchNorm & 89.06 & -0.27\\
                ER & 89.34 & +0.02\\
                Aug & 89.72 & +0.40\\
                Per-Entity Learnable Offset & 89.91 & +0.59\\
                Per-Joint Learnable Offset & 89.46 & +0.14\\
                Global Learnable Offset & 90.02 & +0.70\\
                S2CoM\dag/STD & 90.29 & +0.97\\
                S2CoM\dag & 90.79 & +1.47\\
            \hline
                \textbf{CHASE (Ours)} & \textbf{91.30} & \textbf{+1.98} \\
            \hline
	\end{tabular}
\end{table}

\subsection{Quantitative Results}

\textbf{Recognition Performance.} Table~\ref{tab:sota} summarizes the experimental results on 7 multi-entity action recognition datasets. For fair comparisons \citep{igformer2022,wen2023interactive}, only the joint modality is used in this experiment. Top-1 accuracy serves as the evaluation metric. The results reveal two key observations: 1) We compare the proposed CHASE framework with 5 baseline single-entity backbone models (indicated by the light blue background). \uline{By employing CHASE to mitigate entity bias in skeletal data, we consistently enhance the performance of all baseline models.} For instance, CHASE significantly improves the state-of-the-art single-entity backbone DeGCN \citep{degcn2024tip}, demonstrating a notable performance boost across various settings. The performance improvements differ across baseline models and benchmarks, primarily due to variations in the extent of entity bias, which is influenced by factors such as skeleton type and backbone architecture. 2) We further compare CHASE-augmented models with several state-of-the-art multi-entity encoders (indicated by the light yellow background), which often feature complex designs and limited applicability. \uline{CHASE enables simpler models employing a late fusion strategy to perform effectively across diverse scenarios, even surpassing these advanced methods.} Notable examples include outperforming ISTA-Net \citep{wen2023interactive}, me-GCN \citep{liu2024learning}, ASEA \citep{chen2025ASEA}, and HandFormer \citep{shamil2024HandFormer}. It suggests that CHASE generalizes well to various recognition tasks, regardless of dataset complexity.

\begin{figure*}[t]
    \begin{center}
    \includegraphics[width=\linewidth]{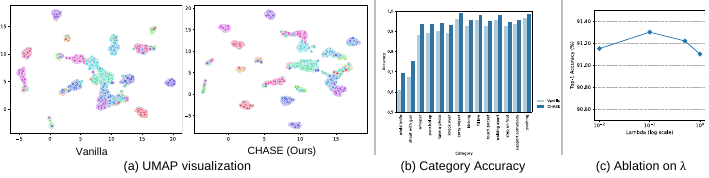}   
    \end{center}
    \vspace{-1.0em}
    \caption{\textbf{More qualitative results}. (a) UMAP \citep{mcinnes2018umap-software} visualizations of skeleton sequence representations on the test split of NTU26 X-Sub. Compared with Vanilla, our proposed CHASE differentiate similar actions and interactions better by assisting backbones to learn more distinctive representations. (b) Category-level accuracy comparison between the vanilla baseline and CHASE. It demonstrates the effectiveness of CHASE by improving recognition accuracy for most categories. (c) Ablation study on \(\lambda\) for MPMMD objective.}
    \label{fig:confusion}
    \vspace{-1.0em}
\end{figure*}

\begin{table*}[t]
	\centering
	\caption{Analysis of Inter-entity Distribution Discrepancies}
	\label{discrepancy}
	\begin{tabular}{l|c|c|c|c|c|c}
            \hline
                Set& Method & Avg KLD \(\downarrow\) & JSD \(\downarrow\) & BD \(\downarrow\) & HD \(\downarrow\) & MMD \(\downarrow\) \\
		\hline
                \multirow{2}{*}{\uppercase\expandafter{\romannumeral1}}& Vanilla & 1.07 & 0.19 & 0.25 & 0.46 & 0.94\\
                & \textbf{CHASE} & \textbf{0.39} & \textbf{0.08} & \textbf{0.10} & \textbf{0.30} & \textbf{0.05} \\
            \hline
                \multirow{2}{*}{\uppercase\expandafter{\romannumeral2}}& Vanilla & 1.00 & 0.18 & 0.23 & 0.45 & 1.03\\
                & \textbf{CHASE} & \textbf{0.45} & \textbf{0.10} & \textbf{0.11} & \textbf{0.32} & \textbf{0.07} \\
            \hline
                \multirow{2}{*}{\uppercase\expandafter{\romannumeral3}}& Vanilla & 0.72 & 0.14 & 0.17 & 0.39 & 1.25\\
                & \textbf{CHASE} & \textbf{0.41} & \textbf{0.08} & \textbf{0.10} & \textbf{0.30} & \textbf{0.05} \\
            \hline
                \multirow{2}{*}{\uppercase\expandafter{\romannumeral4}}& Vanilla & 0.75 & 0.14 & 0.17 & 0.40 & 1.15\\
                & \textbf{CHASE} & \textbf{0.41} & \textbf{0.08} & \textbf{0.09} & \textbf{0.30} & \textbf{0.04} \\
            \hline    
	\end{tabular}
\end{table*}

\begin{table*}[t]
	\centering
	\caption{Analysis of Key Components in CHASE. * indicates non-convergence due to large search space of shift vector.}
	\label{ablation:key}
	       \begin{tabular}{c|c|c|c|c|c|c}
            \hline
                \multicolumn{3}{c|}{CHAS}&
                \multirow{2}{*}{MPMMD}&
                \multirow{2}{*}{LR}&
                \multirow{2}{*}{Acc (\%)} & \multirow{2}{*}{\(\Delta\) (\%)}\\
            \cline{1-3}
                AS & CHC & CLB &  &  & &\\
            \hline
                \Checkmark & \Checkmark & \Checkmark & \Checkmark & 0.1 & 91.30 & - \\
            \hline
                \Checkmark &  & \Checkmark & \Checkmark & 0.1 & 22.65* & -68.65\\
                \Checkmark &  & \Checkmark & \Checkmark & 0.01 & 86.99 & -4.32 \\
                \Checkmark & \Checkmark &  & \Checkmark & 0.1 & 91.20 & -0.10 \\
                \Checkmark &  &  & \Checkmark & 0.1 & 22.75* & -68.56 \\
                \Checkmark &  &  & \Checkmark & 0.01 & 23.51* & -67.79\\
                 & \Checkmark & \Checkmark & \Checkmark & 0.1 & 20.42* & -70.88 \\
                \Checkmark & \Checkmark & \Checkmark &  & 0.1 & 91.17 & -0.13\\
                 &  &  &  & 0.1 & 89.50 & -1.81 \\
            \hline
	       \end{tabular}
\end{table*}

\textbf{Comparison with Alternatives.} We evaluate our proposed CHASE framework against several alternative methods: 1) Vanilla: Directly utilize raw world or pixel coordinates. 2) S2CoM: Relocate each local origin to the spatiotemporal center of mass for individual entities. 3) BatchNorm \citep{bn2015}: Introduce an additional Batch Normalization step immediately after batches of samples are fed into the model. 4) ER (Entity Rearrangement) \citep{wen2023interactive}: A strategy designed to disrupt the inherent order of entities to better model interactions. 5) Aug: Enhance data through random shifts applied to skeleton sequences as an augmentation technique. 6) Per-Entity Learnable Offset: Each entity is assigned a dedicated learnable 3D offset vector that is subtracted from all its joints, directly targeting entity-level bias with static learned parameters. 7) Per-Joint Learnable Offset: Each joint is assigned a learnable 3D offset applied uniformly across all samples, targeting joint-level spatial bias. 8) Global Learnable Offset: A single learnable 3D offset applied to all samples uniformly, representing the simplest form of learnable bias correction. 9) S2CoM\dag: Reposition the global origin to the overall spatiotemporal center of mass. 10) S2CoM\dag/STD: Normalize data by scaling channels based on their standard deviations after applying S2CoM\dag. Table~\ref{ablation:alternatives} demonstrates that \uline{CHASE consistently achieves the highest accuracy improvements among all these methods}. Notably, while the three learnable offset baselines (Per-Entity, Per-Joint, Global) yield modest improvements over vanilla, they fall substantially short of CHASE. This demonstrates that CHASE's gains come from the sample-adaptive shift mechanism and the convex hull constraint, rather than simply learning bias correction parameters: static offsets per entity or joint cannot capture the within-class diversity of entity configurations that CHASE handles dynamically for each input sample.

\textbf{Entity Bias.} Table~\ref{discrepancy} presents metrics evaluating the entity bias on the test sets \uppercase\expandafter{\romannumeral1}-\uppercase\expandafter{\romannumeral4} following \citep{wen2024chase}'s settings. We measure the pair-wise distributions of sampled data points from different entities using \textit{Averaged Kullback-Leibler Divergence} (Avg KLD), \textit{Jensen-Shannon Divergence} (JSD), \textit{Bhattacharyya Distance} (BD), \textit{Hellinger Distance} (HD) and \textit{MMD}. Table~\ref{discrepancy} demonstrates that \uline{CHASE significantly minimizes discrepancies across all evaluation metrics, thereby benefiting backbone learning for each entity}.

\subsection{Qualitative Results}

Fig.~\ref{fig:vizchas} illustrates how CHASE effectively mitigates the entity bias across various data scales. By incorporating CHASE, entity distributions become more aligned in terms of both mean and covariance. Additionally, Fig.~\ref{fig:confusion} (a) compares the UMAP visualization of representations learned by the vanilla CTR-GCN and its CHASE-enhanced counterpart. It demonstrates that CHASE enables the backbone to learn more distinctive feature representations, further enhancing its ability to differentiate between similar categories. Furthermore, category-level accuracy scores, presented in Fig.~\ref{fig:confusion} (b), further highlight that CHASE improves the backbone’s ability to recognize complex actions. For instance, actions such as \textit{wield knife} (+8.50\%), \textit{shoot with gun} (+8.00\%), \textit{whisper} (+5.39\%), and \textit{punch/slap} (+4.37\%) show significant gains with CHASE integration.

\subsection{Ablation Study}

\begin{table*}[t]
	\centering
	\caption{Sub-Entity Strategy on Single-Entity Actions}
	\label{tab:sota_single}
	\begin{tabular}{l|c|c|c|c}
		\hline
		  \multicolumn{1}{l|}{\multirow{2}{*}{Method}}&
            \multirow{2}{*}{Setting*}&
            \multirow{2}{*}{\# Param.}&
            \multicolumn{2}{c}{NTU94(\%)}
            \\
            \cline{4-5}
		&
            &
            &
            X-Sub&
            X-Set\\
        \hline 
            CTR-GCN&
            Full Body&
            1.46M&
            82.69\({}_{(\pm 0.10)}\)&
            85.04\({}_{(\pm 0.04)}\)\\
        \hline 
            CTR-GCN&
            2 Parts&
            1.44M&
            83.22\({}_{(\pm 0.13)}\)&
            85.81\({}_{(\pm 0.03)}\)\\
            \textbf{+ CHASE}&
            2 Parts&
            1.46M&
            \textbf{83.60\({}_{(\pm 0.08)}\)}&
            \textbf{85.85\({}_{(\pm 0.01)}\)}\\
        \hline 
            CTR-GCN&
            5 Parts&
            1.44M&
            81.70\({}_{(\pm 0.10)}\)&
            84.08\({}_{(\pm 0.05)}\)\\
            \textbf{+ CHASE}&
            5 Parts&
            1.45M&
            \textbf{81.99\({}_{(\pm 0.07)}\)}&
            \textbf{84.44\({}_{(\pm 0.03)}\)}\\
		\hline
	\end{tabular}
\end{table*}

In this section, we investigate the effectiveness and efficiency of CHASE by answering a series of questions with ablation studies. Unless specified otherwise, we report the experimental results on NTU26 X-Sub \citep{NTU120} criterion using CTR-GCN \citep{CTR-GCN2021} as the baseline model, which is a well-established evaluation protocol for ablation \citep{igformer2022,wen2023interactive,liu2024learning}.

\textbf{Do all key components contribute to overall performance?} The effectiveness of each component is analyzed in Table~\ref{ablation:key}. Removing the convex hull constraint (CHC) leads to a substantial accuracy reduction of over 60\% at initial learning rates of 0.1 and 0.01. This sharp decline underscores the pivotal role of CHC in facilitating adaptive shift learning. Furthermore, substituting Adaptive Shift (AS) with \(\hat{X}=XWJ_{1,U}\) significantly impairs accuracy, revealing that merely introducing a comparable number of trainable parameters without incorporating an adaptive shift mechanism is insufficient. Lastly, Table~\ref{ablation:key} highlights that CHASE also gains improvements from the inclusion of CLB and MPMMD.

\begin{table}[t]
	\centering
	\caption{Intra-Skeleton Modalities and Ensemble}
	\label{ablation:modal}
	\begin{tabular}{l|c|c}
            \hline
                Modality & Method & Acc (\%) \\
		  \hline
                \multirow{2}{*}{Joint} & Vanilla & 87.07 \\
                & \textbf{CHASE} & \textbf{87.42} \\
            \hline
                \multirow{2}{*}{Bone} & Vanilla & 88.31\\
                 & \textbf{CHASE} & \textbf{88.81} \\
            \hline
                \multirow{2}{*}{Joint Velocity} & Vanilla & 83.38\\
                 & \textbf{CHASE} & \textbf{83.50} \\
            \hline
                \multirow{2}{*}{JBF} & Vanilla & 89.33 \\
                 & \textbf{CHASE} & \textbf{89.65} \\
            \hline
                \multirow{2}{*}{Ensemble} & Vanilla & 90.55 \\
                 & \textbf{CHASE} & \textbf{90.86} \\
            \hline
	\end{tabular}
\end{table}

\textbf{Is the sub-entity strategy effective for adapting CHASE to single-entity skeletons?} 
Table~\ref{tab:sota_single} reports the results on NTU94, a subset of NTU120 focusing on individual actions. We evaluate three settings: Full Body (F.B.), 2 Parts (2.P.), and 5 Parts (5.P.), with the latter two representing possible implementations of the sub-entity strategy. The results demonstrate that selecting a reasonable approach to define sub-entities, such as 2.P., leads to improvements in both accuracy and model efficiency. The improvements remain consistent across independent training runs with different seed initializations, as confirmed by the standard deviations reported in Table~\ref{tab:sota_single}. Furthermore, incorporating CHASE into the baseline models further enhances recognition accuracy while introducing only a minimal increase in model size. We note that the performance gains in the single-entity setting are more modest compared to multi-entity scenarios. This is expected: in the multi-entity setting, different persons occupy distinct spatial locations, resulting in large inter-entity distribution discrepancies that CHASE can effectively reduce. In the single-entity setting under the sub-entity strategy, the \textit{entities} are body parts of the same person, whose spatial extents are much closer to each other within the same skeleton. The inherently smaller distribution gap among sub-entities limits the room for de-biasing, leading to smaller but still consistent improvements.

To verify that distributional discrepancies exist among sub-entities and that CHASE reduces them, we measure pairwise distribution distances between sub-entities on NTU94, following the same protocol as Table~\ref{discrepancy}. We use the 5-part decomposition for this diagnostic analysis because it yields C(5,2)\,=\,10 pairs per sample, providing a more statistically stable estimate than the 2-part setting which produces only a single pair. As shown in Table~\ref{ablation:distribution_single}, CHASE consistently reduces all five metrics compared to the Vanilla baseline, confirming that intra-body distributional asymmetries are present and that CHASE mitigates them through the same coordinate-shifting mechanism as in the multi-entity setting. The absolute reductions are smaller than those in Table~\ref{discrepancy} because the inter-sub-entity gap has a different composition from the inter-person gap. In the multi-entity setting, the gap is dominated by coordinate origin offsets between different persons and is largely removable by coordinate normalization. Among sub-entities of the same skeleton, the gap also includes an intrinsic structural component arising from anatomical differences between body parts, such as the distinct motion patterns of arms and legs, which coordinate normalization cannot eliminate. CHASE addresses the removable coordinate origin portion in both cases, and the residual gap in Table~\ref{ablation:distribution_single} reflects this irreducible structural heterogeneity.

\begin{table*}[t]
    \centering
    \caption{Inter-Sub-Entity Distribution Discrepancies on NTU94}
    \label{ablation:distribution_single}
    \begin{tabular}{l|c|c|c|c|c}
    \hline
    Method & Avg KLD $\downarrow$ & JSD $\downarrow$ & BD $\downarrow$ & HD $\downarrow$ & MMD $\downarrow$ \\
    \hline
    Vanilla & 1.898 & 0.300 & 0.473 & 0.567 & 0.914 \\
    \textbf{CHASE} & \textbf{1.887} & \textbf{0.276} & \textbf{0.416} & \textbf{0.545} & \textbf{0.901} \\
    \hline
    \end{tabular}
\end{table*}

\begin{table*}[t]
	\centering
	\caption{Number of Learnable Parameters}
	\label{table:param}
	\begin{tabular}{l|c|c|c|c|c}
            \hline
                 & CTR-GCN & InfoGCN & STSA-Net & HD-GCN & DeGCN\\
            \hline
                Vanilla & 1.44M & 1.54M & 4.13M & 1.65M & 1.36M\\
                CHASE & 1.46M & 1.57M & 4.16M & 1.68M & 1.39M\\ 
                \(\Delta\) & +1.83\% & +1.96\% & +0.60\% & +1.60\% & +1.93\%\\
            \hline
	\end{tabular}
\end{table*}

\textbf{Can CHASE improve models with other intra-skeleton modalities as input?} We verify this on NTU120 X-Sub benchmark by training the state-of-the-art model, DeGCN \citep{degcn2024tip}, using different skeletal modalities as input, followed by obtaining the ensemble result. As shown in Table~\ref{ablation:modal}, CHASE-wrapped encoders consistently achieve better recognition accuracies compared to their vanilla counterparts across various intra-skeleton modalities, including joints (+0.35\%), bones (+0.50\%), joint velocities (+0.12\%), and JBF \citep{degcn2024tip} (+0.32\%). Notably, CHASE also improves the ensemble result by 0.31\%, underscoring its versatility and effectiveness in enhancing recognition performance across multiple modalities.

The varying magnitude of improvements across modalities reflects how entity bias manifests differently in different representations. Bones and JBF, which encode absolute structural relationships directly dependent on coordinate origin choices, benefit more substantially from adaptive coordinate shifting. Joints also demonstrate clear benefits as they represent absolute spatial positions. In contrast, joint velocities, which capture relative motion between consecutive frames, experience smaller performance gains because this modality is less directly affected by static coordinate offsets. These results demonstrate that CHASE's de-biasing strategy effectively addresses entity bias in a manner proportional to each modality's vulnerability to coordinate system effects.

\textbf{Is CHASE lightweight and efficient?} CHASE is designed as a flexible wrapper compatible with various backbones. As shown in Table~\ref{table:param}, it introduces approximately 26.37k trainable parameters, amounting to only a 1\%-2\% increase over the backbone's original parameters, depending on the base model size. The total trainable parameters can be estimated as \((U+1+C_2)\times C_1 + C_2 \times U\). In terms of computational cost, CHASE adds around 2.50M FLOPs. These results confirm that it achieves a balance between effective and lightweight design, making it well-suited for enhancing skeleton-based learning.

\begin{table*}[t]
    \renewcommand\arraystretch{1.3}
	\centering
	\caption{Mixed Recognition of Single- \& Multi-Entity Actions. * indicates this model is implemented and trained using the public code.}
	\label{tab:sota_ntu}
	\begin{tabular}{l|c|c|c|c|c}
		\hline
		  \multicolumn{1}{l|}{\multirow{2}{*}{Method}}&
            \multirow{2}{*}{Venue}&
            \multicolumn{2}{c|}{NTU120(\%)}&
            \multicolumn{2}{c}{NTU60(\%)}
            \\
            \cline{3-6}

		      &
            &
            X-Sub&
            X-Set&
            X-Sub&
            X-View\\
        \hline

            CTR-GCN&
            ICCV'21&
            88.90&
            90.60&
            92.40&
            96.80\\

            InfoGCN&
            CVPR'22&
            89.80&
            91.20&
            93.00&
            97.10\\

            STSA-Net&
            Neuro.'23&
            88.50&
            90.70&
            92.70&
            96.70\\

            HD-GCN&
            ICCV'23&
            90.10&
            91.60&
            93.40&
            97.20\\

            SkateFormer&
            ECCV'24&
            89.80&
            91.40&
            93.50&
            97.80\\

            BlockGCN&
            CVPR'24&
            90.30&
            91.50&
            93.10&
            97.00\\
            	
            MMCL&
            MM'24&
            90.30&
            91.70&
            93.50&
            97.40\\

            Shap-Mix&
            IJCAI'24&
            90.40&
            91.70&
            93.70&
            97.10\\

        \hline   

            DeGCN*&
            TIP'24&
            90.55&
            91.83&
            93.32&
            97.14\\

            \textbf{+ CHASE}&
            -&
            \textbf{90.86}&
            \textbf{91.89}&
            \textbf{93.39}&
            \textbf{97.20}\\
		
		\hline
	\end{tabular}
\end{table*}

\textbf{Can CHASE improve performance in mixed single- and multi-entity action recognition?} Table~\ref{tab:sota_ntu} presents the recognition performance for mixed single- and multi-entity actions across the entire NTU120 and NTU60 datasets. The results highlight the comparative effectiveness of the proposed CHASE framework when integrated with the state-of-the-art DeGCN \citep{degcn2024tip}. CHASE-augmented DeGCN achieves competitive or superior performance compared to other related methods, outperforming MMCL \citep{liu2024mmcl} and Shap-Mix \citep{zhang2024shapmix} on more challenging NTU120 benchmark.

\textbf{What are the optimal hyperparameters for MPMMD?} Fig.~\ref{fig:confusion} (c) illustrates the impact of varying the trade-off weight factor \(\lambda\). The best performance is achieved when \(\lambda = 0.1\), indicating that MPMMD functions as an auxiliary objective complementing the primary recognition task objective. Additionally, we experimented with different values of \(M\) but observed insignificant performance differences. Therefore, we set \(M=1\) to ensure computational efficiency.

\begin{table*}[t]
	\centering
	\caption{Test-Time Skeleton Noises and Masking}
	\label{sup:noise}
	\begin{tabular}{l|c|c|c|c}
            \hline
                \multirow{2}{*}{Test-Time} & \multicolumn{2}{c|}{+ Noise (\%)} & \multicolumn{2}{c}{+ Mask (\%)}\\
                \cline{2-5}
                 & \(\sigma=10^{-3}\) & \(\sigma=10^{-2}\) & \(p_m=10^{-2}\) & \(p_m=10^{-1}\)\\
		\hline
                CTR-GCN & 88.55 & 80.72 & 81.15 & 56.37 \\
                \textbf{+ CHASE} & \textbf{91.24} & \textbf{82.53} & \textbf{88.57} & \textbf{60.65} \\
            \hline   
	\end{tabular}
\end{table*}
\textbf{Is the CHASE-wrapped model robust to test-time noise and occlusions?} To assess robustness, we introduce intentional corruption to skeleton sequences during inference, simulating potential occlusions or errors in skeleton estimation. The added noise \(X_n \sim \mathcal{N}(\mu, \sigma^2)\) follows a normal distribution with a mean of \(\mu=0\) and standard deviations of \(\sigma=10^{-3}\) and \(10^{-2}\). For masking, skeleton sequences are randomly masked with probabilities \(p_m=10^{-2}\) and \(10^{-1}\). When tested on noisy and occluded inputs in Table~\ref{sup:noise}, CTR-GCN integrated with CHASE delivers better performance compared to the vanilla version, demonstrating enhanced robustness.

\section{Conclusion}\label{sec:conclusion}

This paper presents CHASE, a normalization method based on Convex Hull Adaptive Shift to minimize Entity bias for skeleton-based action and interaction recognition. To the best of our knowledge, our proposed approach is the first to address the observed entity bias in various skeleton sequences. Our core idea to de-bias is adaptively re-centering the world origin of each sample, thereby unbiasing the subsequent backbones and boosting their performance. CHASE works as a plug-and-play normalization module to reduce entity bias and helps the subsequent classifier gain improved recognition performances across diverse settings. For evaluations, we conduct extensive experiments using 5 baseline backbones on 7 datasets across various numbers and types of entities. The experimental results consistently indicate that CHASE can help reach state-of-the-art performances across a variety of action and interaction recognition tasks, contributing to developing efficient and effective skeleton-based learning models.

\backmatter

\section*{Declarations}

\subsection{Funding}
This work was supported by National Natural Science Foundation of China (No. 62473007), Guangdong Outstanding Youth Fund (No. 2026B1515020015), Shenzhen Innovation in Science and Technology Foundation for The Excellent Youth Scholars (No. RCYX20231211090248064), Southern Marine Science and Engineering Guangdong Laboratory (Zhuhai) (SML2024SP007) and Research Project (No. HDHDW59B010202).

\subsection{Competing Interests}
The authors have no competing interests to declare that are relevant to the content of this article.

\subsection{Data Availability}

The authors declare that the datasets used in this paper are available at the following links:
\begin{enumerate}
    \item NTU RGB+D \citep{NTU60} and NTU RGB+D 120 \citep{NTU120}: \url{https://rose1.ntu.edu.sg/dataset/actionRecognition/}
    \item H2O \citep{H2O_TA-GCN2021}: \url{https://h2odataset.ethz.ch/}
    \item Assembly101 \citep{Assembly101}: \url{https://assembly-101.github.io/}
    \item Collective Activity Dataset \citep{cad2009}: \url{https://cvgl.stanford.edu/projects/collective/collectiveActivity.html}
    \item Volleyball Dataset \citep{msibrahi2016VOL}: \url{https://github.com/mostafa-saad/deep-activity-rec}
    \item HARPER \citep{HARPER2024}: \url{https://github.com/intelligolabs/HARPER}
\end{enumerate}

\bibliography{sn-bibliography}

\end{document}